\documentclass[10pt,twocolumn,letterpaper]{article}

\usepackage[pagenumbers]{cvpr} 
\usepackage{xcolor}

\usepackage{graphicx}
\usepackage{tabularx}
\usepackage{booktabs}

\definecolor{cvprblue}{rgb}{0.21,0.49,0.74}
\newcommand{\vct}[1]{\boldsymbol{#1}} 

\newcommand{\vc}{\vct{c}}
\newcommand{\vf}{\vct{f}}

\newcommand{\vh}{\vct{h}}

\newcommand{\vq}{\vct{q}}

\newcommand{\vx}{{\vct{x}}}

\def\calL{\mathcal{L}}

\def\calX{\mathcal{X}}

\newcommand{\expec}{\mathbb{E}}

\usepackage[pagebackref,breaklinks,colorlinks,citecolor=cvprblue]{hyperref}

\def\paperID{5754} 
\def\confName{CVPR}
\def\confYear{2024}

\title{ImageDream: Image-Prompt Multi-view Diffusion for 3D Generation}

\author{\href{https://pengwangucla.github.io/peng-wang.github.io/}{Peng Wang} ~~~~~~~~~~~~~~ \href{https://seasonsh.github.io/}{Yichun Shi} \\
ByteDance, USA \\
\tt\small \{peng.wang, yichun.shi\}@bytedance.com}

\begin{document}

\twocolumn[{
      \vspace{-1em}
      \maketitle
      \vspace{-1em}
      \begin{center}
        \centering
        \vspace{-0.2in}
        \includegraphics[width=1.03\linewidth]{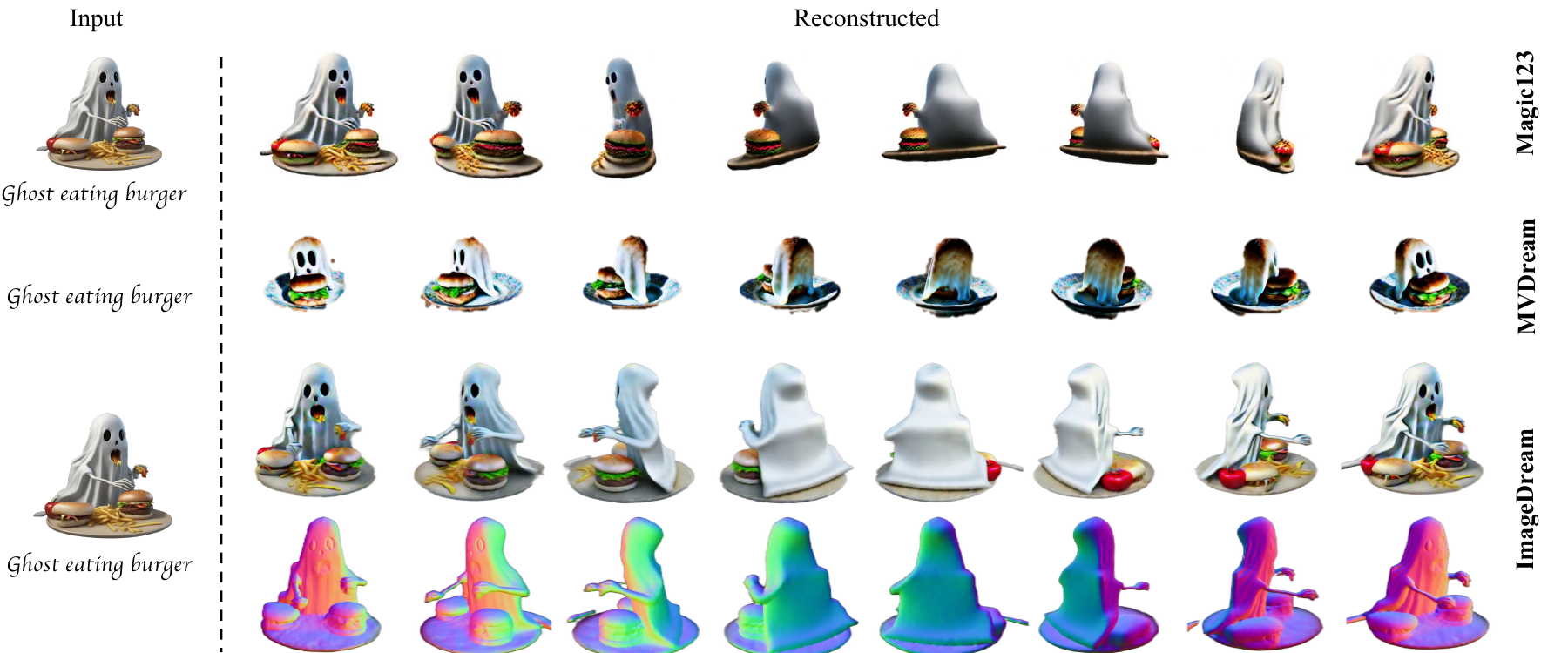}
        \vspace{-0.2in}
        \captionof{figure}{
          \textbf{ImageDream} is a novel framework that generates high quality 3D model from any viewpoint given a single image. It vastly improves the 3D geometry quality comparing to previous SoTA, e.g. Magic123~\cite{qian2023magic123}, and more importantly, it inherits the great text image alignment from the generated image-prompt, comparing with MVDream~\cite{shi2023MVDream}. Here, we provide 8 views of a generated object from different methods, and in the last row, we show the corresponding normal maps rendered with ImageDream generated model. 
        }
        \label{fig:teaser}
      \end{center}
    }]

\maketitle

\begin{abstract}
We introduce \textit{ImageDream}, a novel Image-Prompt Multi-view diffusion model devised for 3D object generation. \textit{ImageDream} excels in delivering 3D models of superior quality comparing with other State-of-the-Art (SoTA) image-conditioned endeavors. Specifically, we consider a canonical camera coordination of the object in image rather than relative. This enhancement significantly augments the visual geometry correctness. Our models are meticulously crafted, taking into account varying degrees of control granularity derived from the provided image: wherein, the global control predominantly influences the object layout, whereas the local control adeptly refines the image appearance. The prowess of \textit{ImageDream} is empirically showcased through a comprehensive evaluation predicated on a common prompt list as delineated in MVDream~\cite{shi2023MVDream}. 
Our project page is \url{https://Image-Dream.github.io}.
\end{abstract}    
\section{Introduction}
\label{sec:intro}

\begin{figure*}[t]
    \centering\vspace{-0.5em}
    \includegraphics[width=0.95\linewidth]{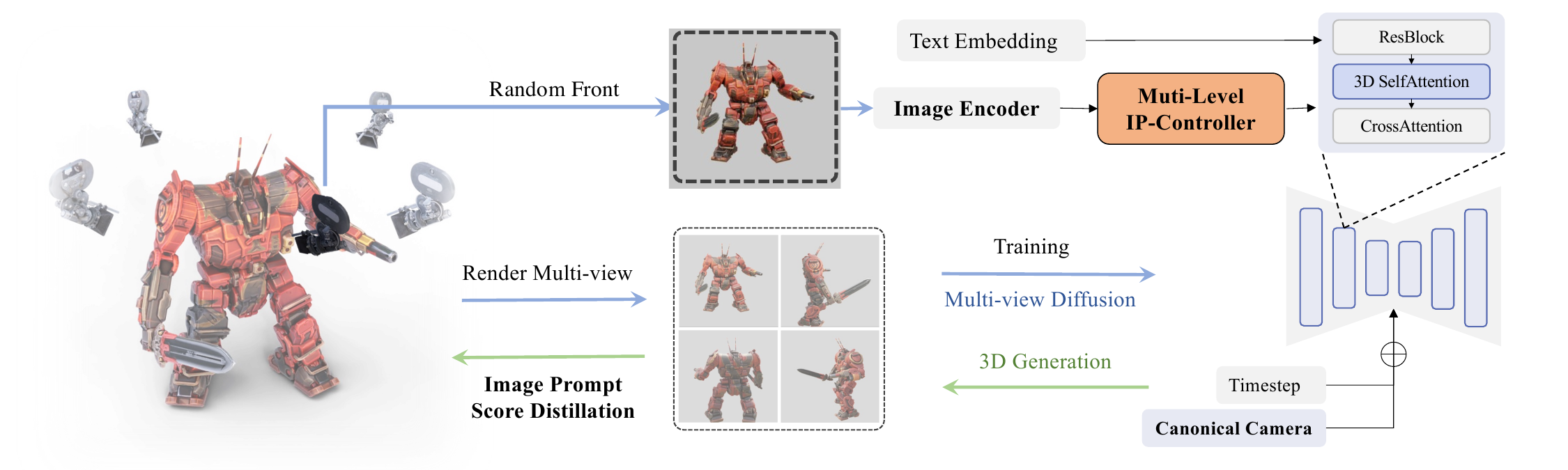}\vspace{-0.5em}
    \caption{The training pipeline of ImageDream. The blue arrow indicates training of the diffusion network and the green arrow indicates training of NeRF model. 
    In diffusion training, given a 3D object, we first render multiple views based on canonical camera coordination (bottom), and render another image-prompt front-view images with a random setting (top). The multi-view images are fed as training targets for multi-view diffusion networks, and image-prompt is encoded with a multi-level controller as input to the diffusion. In NeRF training, we use the trained diffusion for image-prompt score distillation. 
    }\vspace{-1.5em}
    \label{fig:framework}
\end{figure*}

In the domain of 3D generation, incorporating images as an additional modality for 3D generation, compared to methods relying solely on text~\cite{poole2022dreamfusion}, offers significant advantages, as the common saying, \textit{An image is worth a thousand words}. Primarily, images convey rich, precise visual information that text might ambiguously describe or entirely omit. 
For instance, subtle details like textures, colors, and spatial relationships can be directly and unambiguously captured in an image, whereas a text description might struggle to convey the same level of detail comprehensively or might require excessively lengthy descriptions. 
This visual specificity aids in generating more accurate and detailed 3D models, as the system can directly reference actual visual cues rather than interpret textual descriptions, which can vary greatly in detail and subjectivity. Moreover, using images allows for a more intuitive and direct way for users to communicate their desired outcomes, particularly for those who may find it challenging to articulate their visions textually. This multimodal approach, combining the richness of visual data with the contextual depth of text, leads to a more robust, user-friendly, and efficient 3D generation process, catering to a wider range of creative and practical applications.

Adopting images as an additional modality for 3D object generation, while beneficial, also introduces several challenges, Unlike text, images contain a multitude of features like color, texture, spatial relationships that are more complex to analyze and interpret accurately with a solely encoder like CLIP~\cite{CLIP}. In addition, high variant of light, shape or self-occlusion of the object can lead to inaccurate and in-consistent view synthesis, therefore leading blurry or incomplete 3D models. 

The complexity of image processing necessitates advanced, computationally intensive algorithms to accurately decode visual information and ensure consistent appearance across multiple views. Researchers have employed various strategies with diffusion models, such as Zero123~\cite{zero123}, and other recent works~\cite{qian2023magic123, liu2023syncdreamer}, to elevate a 2D object image to a 3D model. However, a limitation of image-only solutions is that, although the synthesized views are visually impressive, the reconstructed models often lack geometric accuracy and detailed textures, particularly in the object's rear views. This issue primarily stems from significant geometric inconsistencies across the generated or synthesized views. Consequently, during reconstruction, non-matching pixels are averaged in the final 3D model, leading to indistinct textures and smoothed geometry.

Fundamentally, image-conditioned 3D generation represents an optimization problem with more stringent constraints compared to text-conditioned generation. Hence, achieving optimized 3D models with clear details is more challenging, as the optimization process is prone to deviating from the trained distributions due to the limited amount of 3D data. For example, generating a horse based solely on text descriptions may yield detailed models if the training dataset includes a variety of horse styles. However, when an image specifies particular textures, shapes, and fur details, the novel-view texture generation may easily deviate from the trained distributions.

In this paper, we introduce ImageDream to address these challenges. Our approach involves considering a canonical camera coordination across different object instances and designing a multi-level image-prompt controller that can be seamlessly integrated into the existing architecture. Specifically, the canonical camera coordination mandates that the rendered image, under default camera settings (i.e., identity rotation and zero translation), represents the object's centered front-view. This significantly simplifies the task of mapping variations in the input image to 3D. The multi-level controller offers hierarchical control, guiding the diffusion model from the image input to each architectural block, thereby streamlining the path of information transfer.

As illustrated in Fig.\ref{fig:teaser}, ImageDream excels in generating objects with correct geometry from a given image, enabling users to leverage well-developed image generation models\cite{podell2023sdxl} for better image-text alignment than purely text-conditioned models like MVDream~\cite{shi2023MVDream}. Furthermore, ImageDream surpasses existing state-of-the-art (SoTA) zero-shot single image 3D model generators, such as Magic123~\cite{qian2023magic123}, in terms of geometry and texture quality. Our comprehensive evaluation in the experimental section (Sec.~\ref{sec:exp}), which includes both qualitative comparisons through user studies and quantitative analyses, demonstrates ImageDream's superiority over other SoTA methods.

\section{Related Works}
\label{sec:related}

\begin{figure*}[t]
    \centering\vspace{-0.5em}
    \includegraphics[width=\linewidth]{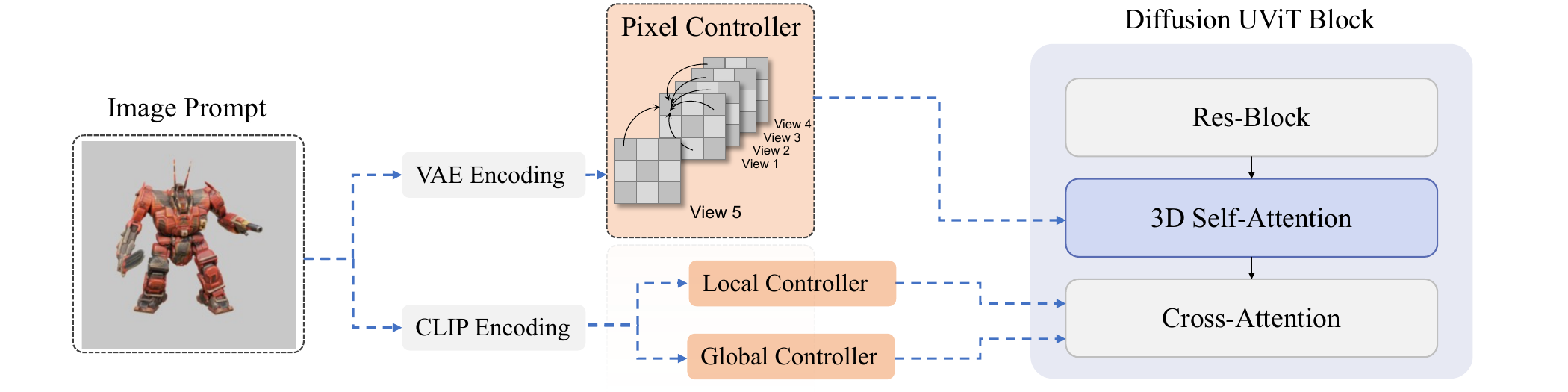}\vspace{-0.5em}
    \caption{The Multi-level Controller of ImageDream. Given an image prompt, global controller and local controller takes input of image features after CLIP encoding, and then output adapted features to cross-attention layers. It represents image semantic information. The pixel controller send the VAE encoded feature to diffusion, and perform pixel-level dense self-attention with corresponding hidden features at each layer of the four-view MVDiffusion.
    }\vspace{-1.5em}
    \label{fig:framework}
\end{figure*}

We recognize that 3D generation is a well-established field; this review focuses on significant advancements closely related to our research.

\noindent\textbf{Text-to-3D Generation with Diffusion.}
The emergence of deep generative models has significantly impacted 3D generation. Early methods targeted the reconstruction of simple objects using multi-view rendered images~\cite{henderson2020learning,henderson2020leveraging}. The evolution of these techniques, from Generative Adversarial Networks (GANs)\cite{nguyen2019hologan,nguyen2020blockgan,niemeyer2021giraffe,deng2022gram,chan2022efficient,gao2022get3d} to diffusion-based frameworks\cite{LatentDiffusion}, marks a notable progression.

Recent 3D diffusion models, specifically for tri-planes~\cite{shue2023triplanediffusion,wang2023rodin} and feature grids~\cite{karnewar2023holodiffusion}, have emerged. However, these models often focus on specific objects like faces and ShapeNet~\cite{chang2015shapenet} objects. Concurrently, there's growing interest in reconstructing object shapes from monocular image inputs~\cite{wu2023mcc,nichol2022point,jun2023shap}, demonstrating the evolving stability of image generation methodologies. A significant challenge remains in generalizing these models to the extent of their 2D counterparts, likely due to constraints in 3D data size, representation, and architectural design.

\noindent\textbf{Lifting 2D Diffusion for 3D Generation.}
In light of the limited generalizability of direct 3D generative models, a parallel line of research has explored the elevation of 2D diffusion priors into 3D generation, often integrating with 3D representations like NeRF~\cite{mildenhall2021nerf}. A pivotal approach in this area is the score distillation sampling (\textbf{SDS}) introduced by Poole et al.\cite{poole2022dreamfusion}, using diffusion priors as score functions to guide 3D representation optimization. Alongside Dreamfusion, works like SJC\cite{wang2023score}, which utilize stable-diffusion models~\cite{LatentDiffusion}, have emerged. Subsequent studies have focused on enhancing 3D representations~\cite{lin2023magic3d, tsalicoglou2023textmesh,tang2023make,chen2023fantasia3d}, refining sampling schedules~\cite{huang2023dreamtime}, and optimizing loss designs~\cite{wang2023prolificdreamer}. Despite their ability to generate photorealistic objects of various types without 3D data training, these methods struggle with multi-view consistency. Moreover, each 3D model requires individualized optimization through prompt and hyper-parameter adjustments. 
Notably, MVDream~\cite{shi2023MVDream} enhances generation robustness by joint training with 2D and 3D datasets, producing satisfactory results with uniform parameters, drawing on multi-view diffusion via SDS. Our work builds upon these concepts, applying them to image-prompt generation and retaining the robustness characteristic of MVDream.

\noindent\textbf{Image-based Novel View Synthesis.}
Direct synthesis of novel 3D views from single images has also been explored, bypassing traditional reconstruction processes. Watson et al.\cite{DBLP:conf/iclr/WatsonCMHT023} pioneered diffusion model applications in view synthesis as the pipeline in Sitzmann et al.\cite{sitzmann2019scene} using the ShapeNet dataset. Subsequent advancements include Zhou et al.'s~\cite{zhou2023sparsefusion} extension to latent space with an epipolar feature transformer and Chan et al.'s~\cite{chan2023genvs} approach to enhance view consistency. Szymanowicz et al.\cite{szymanowicz2023viewset} proposed a multi-view reconstructor using unprojected feature grids. A common limitation across these methods is their dependency on specific training data, with no established adaptability to diverse image inputs. Fine-tuning pre-trained image diffusion models\cite{sdvariation} on extensive 3D render datasets for novel view synthesis, as proposed by Zero123~\cite{zero123}, remains constrained by geometric consistency issues. Later works, including SyncDreamer~\cite{liu2023syncdreamer}, Consistent 1-to-3~\cite{ye2023consistent}, and Zero123plus~\cite{shi2023zero123++}, have sought to enhance multi-view consistency through joint diffusion processes, but the reconstruction of geometrically coherent 3D models remains a challenge.

\noindent\textbf{Single Image-conditioned Reconstruction.}
Recent advances in deriving 3D models from single or few images predominantly leverage NeRF representations. Techniques such as RegNeRF~\cite{DBLP:conf/cvpr/NiemeyerBMS0R22}, which uses geometry loss from depth patches, and SinNeRF~\cite{Xu_2022_SinNeRF}, RealFusion~\cite{melas2023realfusion}, and NeuralLift~\cite{Xu_2022_neuralLift}, which combine depth maps or Score Distillation Sampling during NeRF training, represent significant steps forward. Despite their effectiveness, the quality of these generated models remains suboptimal for real-world applications. Magic123~\cite{qian2023magic123} combines single-view and novel-view diffusion networks, achieving impressive texture quality in 3D models. However, our tests reveal limitations in understanding correct object geometry.

We also note recent parallel developments, such as Wonder3D~\cite{long2023wonder3d}, which incorporate normal diffused outputs into original diffusion models, and DreamCraft3D~\cite{sun2023dreamcraft3d}, which employ a second-stage DreamBooth-like model fine-tuning for enhanced texture modeling. These works, while promising, remain distinct from our contributions.

\section{Methodology}

In this section, we first talk about the MVDream~\cite{shi2023MVDream} pipeline and then describe our method to input the image prompt. 

\begin{figure*}[t]
    \centering\vspace{-0.5em}
    \includegraphics[width=\linewidth]{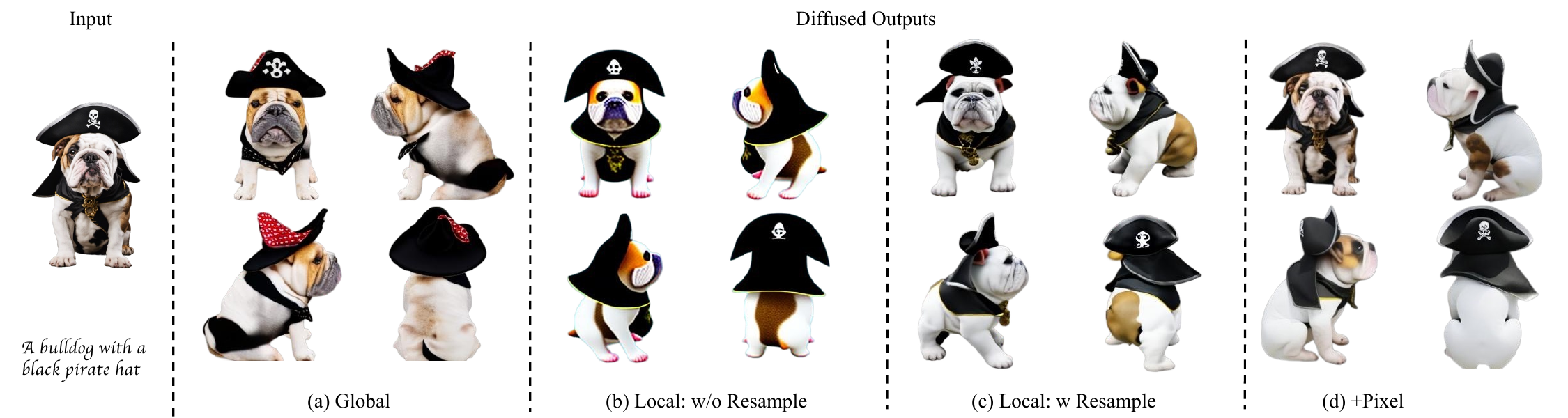}\vspace{-0.5em}
    \caption{An example of the diffused results from different settings of multi-level controllers in ImageDream (see Sec.~\ref{subsec:multi-level}).
    }\vspace{-1.5em}
    \label{fig:control}
\end{figure*}

\subsection{Preliminary}
\label{subsec:pre}
In MVDream, there are two stage for 3D model production. The first stage is training a multi-view diffusion network that produces four orthogonal and consistent multi-view images from a text-prompt given respective camera embedding. 
In the second stage, a multi-view score distillation sampling (MV-SDS) is adopted to produce a detailed 3D NeRF model.

In the first stage, each block of the multi-view network contains a densely connected 3D attention on the four view images, which allows a strong interaction in learning the correspondence relationship between different views. To train such a network,
it adopts a joint training with the rendered dataset from the Objaverse~\cite{deitke2023objaverse} and a larger scale text-to-image (t2i) dataset, LAION5B~\cite{schuhmann2022laion}, to maintain the generalizability of the fine-tuned model.  
Formally, given text-image dataset $\calX = \{\vx, y\}$ and a multi-view dataset $\calX_{mv} = \{\vx_{mv}, y, \vc_{mv}\} $, where $\vx$ is an latent image embedding from VAE~\cite{kingma2013auto}, $y$ is a text embedding from CLIP~\cite{CLIP}, and $\vc$ is their self-desgined camera embedding, we may formulate the the multi-view (MV) diffusion loss as,
{
\small
\begin{align}
\calL_{\text{MV}}(\theta, \calX, \calX_{mv}) &= \expec_{\vx, y, \mathbf{c}, t, \epsilon}\left[ \Vert \epsilon - \epsilon_\theta(\vx^p; y,\vc^p,t) \Vert_{2}^{2} \right]  \label{eq:diffuse} \\
\text{where,~} (\vx^p, \vc^p) &= \begin{cases} 
(\vx, \textbf{0}) & \text{with probability } p  \\
(\vx_{mv}, \vc_{mv}) & \text{with probability } 1-p 
\end{cases} \nonumber 
\end{align}
}
here, $\vx$ is the noisy latent image generated from a random noise $\epsilon$ and image latent, the $\epsilon_\theta$ is the multi-view diffusion (MVDiffusion) model parametrized by $\theta$.

After the model is trained, the MVDiffusion model can be inserted to the DreamFusion~\cite{poole2022dreamfusion} pipeline, where the authors adopt a score-distillation sampling (SDS) based on the four generated views. Specifically, in each iteration step, a random 4 orthogonal views are rendered from a NeRF $g(\phi)$ with a random 4 view camera extrinsic and intrinsic $\vc$.  Then, they are encoded to latents $\vx^{mv}$ and inserted to the multi view diffusion network to compute a diffusion loss in the image space which is back propagated to optimize the NeRF parameters.   Formally, 
 \begin{equation}
     \calL_{\text{MV-SDS}}(\phi,\vx^{mv}) = \expec_{t,\mathbf{c},\epsilon}\Big[\Vert \vx^{mv} - \hat{\vx}^{mv}_0 \Vert_{2}^2 \Big].
    \label{eq:x0_sds}
 \end{equation}
Here, $\hat{\vx}^{mv}_0$ s the denoised MV image at timestep $0$ from MVDiffusion. After fusion, MVDream shows significant improvement of object geometry correctness without the Janus issues.

\subsection{Canonical Camera}
In the context of MVDream, a critical observation is the diffusion of multi-view images using a global aligned camera coordination. In other words, the image from a default camera (no azimuth rotation) is always the front view of the object. This is done by asking the CLIP image feature of a view best match the "front view" CLIP text feature embeddings. This alignment facilitates the fusion of diffused images in the fusion step, reducing ambiguity regarding their viewpoints in relation to the provided text prompt.

As emphasized in the introduction, this alignment also reduced the difficulties in learning the accurately reconstructing the geometry of objects. Thereby, in image prompt cases, in contrast to previous image-conditioned approaches like Zero123~\cite{zero123}, which attempt to recover object 3D geometry based on image camera coordination system,  ImageDream adopts canonical/world camera coordination as in MVDream. Our diffusion model aims to regress towards the canonical multiple view image of the object as depicted in the image. This approach is expected to yield superior geometric accuracy compared to systems that utilize relative camera coordination.

Formally, for an image of an object rendered from a random viewpoint with a random camera, denoted as $\vx_r$, we create the ImageDream diffusion multi-view (MV) dataset as $\calX_{mv} = \{\vx_{mv}, y, \vx_r, \vc_{mv}\}$, where $\vc_{mv}$ is the introduced canonical cameras in MVDream. Then, the rest of diffusion loss is the same as Eqn.(\ref{eq:diffuse}).  

\subsection{Multi-level Controllers}
\label{subsec:multi-level}
In order to insert the image prompt to control the output MV images, we consider a multi-level strategy. The overall structure of the multi-level controller from an image prompt can be seen in Fig.~\ref{fig:control}, and we elaborate the details of each component in the following. 

\noindent\textbf{Global Controller.}  In our initial approach, we integrated global CLIP image features into MVDream, akin to how text features are used, by fine-tuning the model's already well-established training. Recognizing that MVDream is primarily trained on text embeddings, we introduced a multi-layer perceptron (MLP) $\theta_g$, functioning as an adaptor similar to IP-Adaptor~\cite{ye2023ip}, following the CLIP image global embedding. This step aims to align image features with text features, ensuring compatibility within the MVDream framework. 
Specifically, CLIP image encoding encodes image feature to a 1024 vector with a token length of 4, which we named as $\vf_g$. And, $\theta_{g}$ further adapts the image feature to be 1024 as the input to cross-attention. 

On the MVDiffusion side, inside of an attention layer $l$, we add a new set of MLPs,  $\theta_{k_g, l}$ and $\theta_{v_g, l}$, that takes the input the adapted features and output its attention key and value matrix, which then aggregated based on the query feature matrix, $\vq_{l}$, yielding a corresponding image cross-attention feature $\vh_{g,l}$. Here, a weight $\lambda = 1.0$ is introduced to balance the hidden from text and image, and the final output of layer $l$ is $\vh_l = \vh_{t,l} + \lambda\vh_{g,l}$. We refer to decoupled cross-attention in IP-Adaptor for additional details.

To train such a model, we freeze the diffusion model, and only fine-tune $\{\theta_{g}, \theta_{k_g,l},  \theta_{v_g,l} \}_l$.  We follow the training setting of MVDream by considering both 3D rendered datasets and 2D image datasets together, which will be elaborated in our experimental section. 

After the model is tuned, we found the model is able of absorb some informations from the image such as structure of the object etc.  As illustrated in  Fig.~\ref{fig:control} (a), comparing with the input, the diffused output is able to put the pirate hat similar to the image on the bulldog, while some detailed pose and appearance information is lost, which we think is not enough for a good control from the input image. 

\noindent\textbf{Local Controller.}  To enhance control, we try to utilize the hidden feature from the CLIP encoder before its global pooling, which likely contains more detailed structural information. This hidden feature, denoted as $\vf_h$, has a token length of $257$ and a feature dimension of $1280$. A MLP adaptor $\theta_{h}$ is introduced to feed $\vf_h$ into the diffusion network's cross-attention module, with $\theta_{k_h,l}$ and $\theta_{v_h,l}$ forming the key and values matrix. These parameters, $\{\theta_{k_h,l}, \theta_{v_h,l}, \theta_{h}\}_l$, are then jointly trained as learnable elements similar to the global controller. Post-training, we observed that the results were overly sensitive to image tokens, leading to overexposed and unrealistic images, especially with higher class free guidance (CFG) settings~\cite{LatentDiffusion}, as shown in Fig.~\ref{fig:control}(b).

To mitigate this, we implemented a resampling module $\theta_r$, following the approach of IP-Adaptor, reducing the hidden token count from 257 to 16, resulting in a more balanced local image feature $\vf_r$. The corresponding local controller parameters are $\{\theta_r, \theta_{k_r,l}, \theta_{v_r,l}\}_l$. As Fig.~\ref{fig:control}(c) illustrates, after this resampling, the diffused images more realistic, even at higher CFG levels.
From the generated images, it's evident that the model captures the overall layout and object shape, but also struggles with finer identity details like object skin texture.


\begin{figure*}[t]
    \centering\vspace{-0.5em}
    \includegraphics[width=0.8\linewidth]{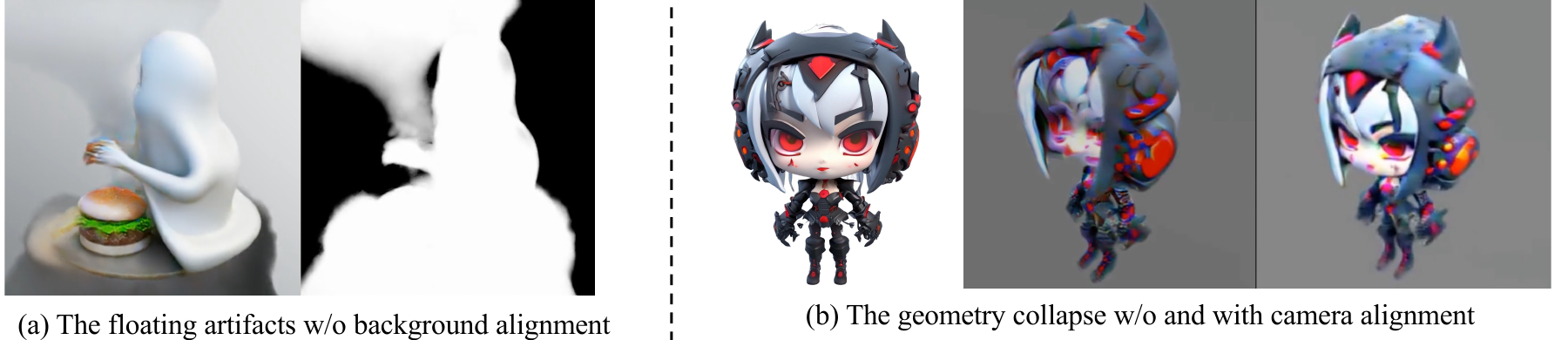}\vspace{-0.5em}
    \caption{Example of artifacts we fixed with image-prompt score distillation (see Sec.~\ref{subsec:sds}). 
    }\vspace{-1.5em}
    \label{fig:artifacts}
\end{figure*}

\noindent\textbf{Pixel Controller.} To optimally integrate object appearance texture, we propose embedding the image prompt pixel latent $\vx$ across all attention layers in ImageDream. Specifically, MVDream employs a 3D dense self-attention mechanism with a shape of $(bz, 4, c, h_l, w_l)$ across four views within a transformer layer. In contrast, ImageDream introduces an additional frame by concatenating the input image, resulting in a feature shape of $(bz, 5, c, h_l, w_l)$. This enables similar 3D self-attention processes between the four-view images and the input image.

During the training of our diffusion network, we refrain from adding noise to the latent from the input image prompt, ensuring the network clearly captures the image information. Additionally, to differentiate the input image features and avoid confusion, we assign an all-zero vector to the camera embedding of the input image. Given that the pixel controller is integrated into the multi-view diffusion without extra parameters, we fine-tune all feature parameters in unison, adopting the same training regime as the global/local controllers but with a learning rate reduced by a factor of ten. This approach preserves the original feature representations more effectively. Post-training, as depicted in Fig~\ref{fig:control}(d), the generated multi-view images not only ethically maintain the appearance from the input image but also uphold the multi-view consistency characteristic of MVDream, resulting in satisfactory 3D model fusion.

Finally, our multi-level controller is a combined one with local and pixel, since we think the global one do not have too much additional information.  There might be potential queries regarding the necessity of a pixel controller, given that IP-Adaptor, relying solely on CLIP features, can capture extensive image texture details: we posit that while IP-Adaptor is effective for modifying objects in the same view as the input, decoding the same view is comparatively simpler. Multi-view diffusion, however, presents a more complex challenge. Decoding from a highly compressed CLIP feature could necessitate prolonged training on larger datasets. Therefore, at this stage, we find the pixel controller significantly beneficial for rapidly training a robust multi-view diffusion model.

\subsection{Image-Prompt Sore Distillation}
\label{subsec:sds}
Implementing the image-prompt multi-view diffusion network in ImageDream follows the multi-view score distillation framework of MVDream (Sec.~\ref{subsec:pre}), with the addition of an image prompt as an input to the diffusion network. However, we need to condier few key differences in NeRF optimization to achieve accurate results.

\noindent\textbf{Background Alignment.} During SDS optimization, the NeRF-rendered image includes a randomly colored background to differentiate the interior and exterior of the 3D object. This random background, when input into the diffusion network alongside the object, can conflict with the background from the image prompt, leading to floating artifacts in the generated NeRF model, as shown in Fig.~\ref{fig:artifacts}(a). To resolve this, we adjusted the image-prompt background to match the rendered background color from NeRF, successfully eliminating these artifacts.

\noindent\textbf{Camera Alignment.} Our diffusion network tends to generate multi-view images mirroring the camera parameters (e.g., elevation, field of view (FoV)) of the input image prompt, parameters which remain unknown during NeRF rendering. Randomly sampling parameters for rendering, as done in MVDream, can result in images incongruent with the image prompt's rendering settings, affecting the geometry of detailed image structures. To mitigate this, we narrowed the parameter sampling range from MVDream's $[15, 60]$, $[0, 30]$ for camera FoV and elevation to $[45, 50]$ and $[0, 5]$, respectively, a range more typical for a generated user photos. This adjustment significantly improved the geometric accuracy of the 3D objects, as demonstrated in Fig.~\ref{fig:artifacts}(b).

We acknowledge this solution's limitations; when the image prompt's camera parameters greatly differ from our selected range in canonical camera setting, the resulting 3D object shape may be unpredictable. Future improvements could include a camera parameter estimation module or increased randomness in the image prompt rendering during diffusion training, to better synchronize the settings between NeRF rendering and diffusion.



\section{Experiments}
\label{sec:exp}

\begin{figure}[t]
    \centering\vspace{-0.5em}
    \includegraphics[width=\linewidth]{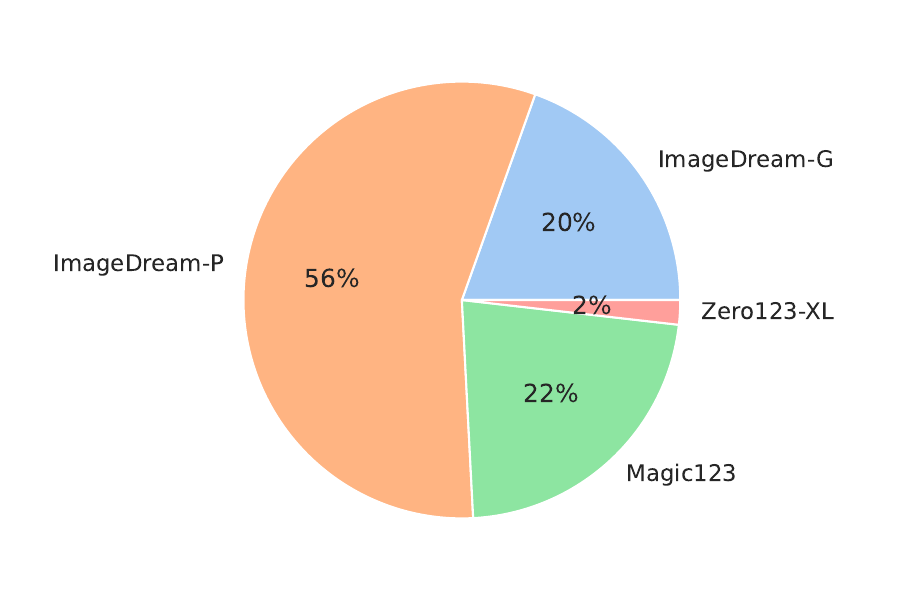}\vspace{-0.5em}
    \caption{User study of different methods. ImageDream-P: our full model w pixel controller, and ImageDream-G: without pixel controller (see Sec.~\ref{subsec:comp}).
    }\vspace{-1.5em}
    \label{fig:user_study}
\end{figure}

\begin{table*}[t]
\captionsetup{font=small} 
\setlength{\tabcolsep}{12pt} 
\renewcommand{\arraystretch}{1.2} 
\fontsize{9}{10}\selectfont 
\begin{center}
\begin{tabularx}{\linewidth}{@{}>{\hsize=7.5\hsize}X|ccc|ccc@{}} 
\toprule
 & \multicolumn{3}{c|}{Synthesized Image}  &  \multicolumn{3}{c}{Re-rendered}  \\
\midrule
Model & QIS(256)$\uparrow$ & CLIP(TX)$\uparrow$ & CLIP(IM)$\uparrow$ & QIS(320)$\uparrow$ & CLIP(TX)$\uparrow$ & CLIP(IM)$\uparrow$ \\
\midrule
SD-XL~\cite{SDXL}    & 52.0 ± 30.5 & 34.6 ± 3.09 & 100 & - & - & - \\
MVDream ~\cite{shi2023MVDream}   & 23.05 ± 14.4 & 31.64 ± 2.99  & 78.41 ± 5.32  & \bf{29.02 ± 10.24} & 32.69 ± 3.39 & 79.63 ± 4.15 \\
\midrule
Zero123~\cite{zero123}    & 22.16 ± 11.16 & 30.42 ± 3.19 & 84.88 ± 5.12  & - & - & - \\ 
Zero123-XL~\cite{zero123} & \bf{34.07 ± 11.64} & 30.80 ± 2.59 & 84.10 ± 4.76 & 28.66 ± 5.03 & 29.19 ± 3.60 & 79.92 ± 6.59 \\
Magic123~\cite{qian2023magic123}    & - & - & - & 24.23 ± 7.68 &29.56 ±4.73  & 82.50 ± 8.78  \\
SyncDreamer~\cite{liu2023syncdreamer}    & 22.04 ± 11.9  & 27.96 ± 3.01 & 78.17 ± 6.13 & 19.84 ± 6.64 & 25.82 ± 3.39 & 73.20 ± 6.30 \\
ImageDream  &  &  &  &  &  &  \\
\; - global & 22.31 ± 7.59 & 32.01 ± 2.84 & 84.50 ± 3.96  & 22.51 ± 5.86 & 31.48 ± 3.32 & 82.58 ± 4.35 \\
\; - local (-G) & 22.49 ± 9.57 & 31.32 ± 2.86 & 82.99 ± 6.03  & 22.30 ± 4.47  & \bf{31.71 ± 2.96}  & 84.34   ±  3.13 \\
\; - +pixel (-P) & 27.10 ± 12.8 & \bf{32.39 ± 2.78}  & \bf{85.69 ± 3.77} & 25.16 ± 6.49  &{31.59 ± 3.23}   & \bf{84.83 ± 4.08} \\
\bottomrule
\end{tabularx}
\caption{Quantitative assessment of image synthesis quality using the test prompt list from MVDream. The DDIM sampler was employed for testing, and Implicit Volume from threestudio was utilized for re-rendering 3D model images. Here Zero123 we took their checkpoint of 165K which including all training instances. 'Zero123-XL' indicates Zero123 trained with the larger Objaverse-xl 10M dataset~\cite{deitke2023objaversexl}, while others were trained solely on the standard Objaverse dataset~\cite{deitke2023objaverse}, which is $\sim$10 times smaller. (-P) and (-G) are correspondent to ImageDream-P and ImageDream-G in Fig.\ref{fig:user_study}.}
\label{tab:synthesis_quality}
\end{center}
\end{table*}

\begin{figure*}[t]
    \centering\vspace{-0.5em}
    \includegraphics[width=\linewidth]{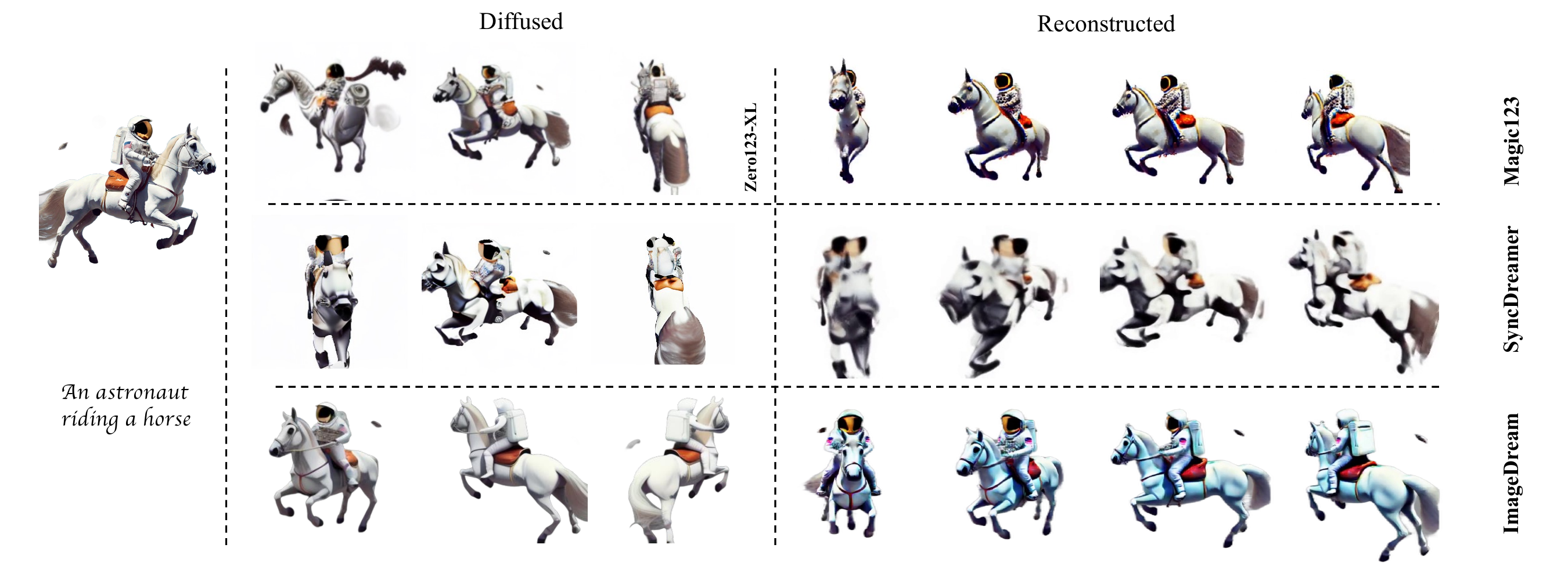}
    \caption{The illustration of synthesized images from different baselines. Diffused: From diffusion models. Reconstructed: Re-rendered images from correspondent fused NeRF model. Please check the webpage for more results.} \vspace{-1.5em}
    \label{fig:quali}
\end{figure*}

In this section, we detail the experimental setup for ImageDream, designed to enable replication of our model. We will release both the model and code following this submission.

\noindent\textbf{Implementation Details.} Adhering to the dataset configuration of MVDream (Sec.~\ref{subsec:multi-level}), we used a combined dataset from Objaverse for 3D multi-view rendering and a 2D image dataset for training controllers (Sec.~\ref{subsec:multi-level}). For image prompts in the 3D dataset, we randomly selected one of the 16 front-side views, with azimuth angles ranging from $[-90, 90]$ degrees, out of the total 32 circle views. For the 2D dataset, we used the input image as the image prompt. A random dropout rate of 0.1 was set for the image prompt during training, replacing dropped prompts with a random uni-colored image. For all experiments, i.e. with global controller, local controller and local plus pixel controllers, we trained for $60$K steps with a batch size of 256 and a gradient accumulation of 2, using the AdamW optimizer. The model is initialized from stable diffusion (2.1) checkpoint of MVDream. The learning rate was set to 1e-4, except for the model with pixel controller, where it was reduced to 1e-5. Test image prompts were resized to $256 \times 256$, and we set the diffusion CFG to 5.0. The training takes $\sim$ 2 days with $8$ A100. For NeRF optimization, we followed MVDream's configuration but introduced a three-stage optimization at resolutions $[64, 192, 256]$ which switched at [$5$K, $10$K] steps, setting the camera distance between $[0.6, 0.85]$ for better NeRF model coverage. The NeRF training is about 1hr with A100. 

\noindent\textbf{Test Dataset.} Our primary focus was evaluating ImageDream outside the Objaverse distribution to ensure its practical applicability. We selected 39 well-curated prompts from MVDream, covering a diverse range of objects with relatively complex geometries and appearances, surpassing datasets like ShapeNet~\cite{chang2015shapenet} or CO3D~\cite{lin2023common}. Using SDXL~\cite{podell2023sdxl}, we generated multiple images from each prompt, selecting ones with aesthetically pleasing objects. The backgrounds of these images were then removed, and the objects re-centered, akin to the approach used in Zero123.

\subsection{Comparisons} 
\label{subsec:comp}
In our evaluation, we compared ImageDream's performance against several SoTA baselines, including Zero123-XL~\cite{zero123} (trained on 10x larger data than ours), Magic123~\cite{qian2023magic123}, and SyncDreamer~\cite{liu2023syncdreamer}. The criteria for comparison were geometry quality and similarity to the image prompt (IP). 'Geometry quality' refers to the generated 3D asset's conformance to common sense in terms of shape and minimal artifacts, while 'similarity to IP' assesses the resemblance of the results to the input image. We executed all baseline tests using default configurations as implemented in threestudio\footnote{\url{https://github.com/threestudio-project/threestudio}}.

\noindent\textbf{Qualitative Evaluation.} Lacking ground truth for the test image prompts, we conducted a real user study to evaluate the quality of the generated 3D models. Participants were briefed on our evaluation standards and asked to choose their preferred model based on these criteria. The experiment was double-blind, with participants shown 3D assets generated by different methods without identifying labels. The comparison results, depicted in Fig.~\ref{fig:user_study}, show that ImageDream, both with (ImageDream-P) and without (ImageDream-G) the pixel controller, significantly outperformed other baselines. ImageDream-P was particularly favored, while ImageDream-G also received a positive preference rate. SyncDreamer was omitted from the figure due to its NeuS results having a 0$\%$ preference rate.

Fig.~\ref{fig:quali} presents a representative case comparing results from the diffusion models and the final NeRF model. Systems like Magic123 and Zero123, which rely on single-view diffusion with relative camera embedding, often produce incorrect geometry, as illustrated by their inability to accurately represent the span of horse body. In contrast, ImageDream, through its unique design, effectively resolves this issue, resulting in more satisfactory models (more results are list in webpage). 


\noindent\textbf{Numerical Evaluation.}
To thoroughly assess image quality at various stages of our generation pipeline, we employed the Inception Score (IS)\cite{InceptionScore} and CLIP scores\cite{CLIP} using text-prompt and image-prompt, respectively. The IS evaluates image quality, while CLIP scores assess text-image and image-image alignment. However, since IS traditionally evaluates both image quality and diversity within a set, and our prompt quantity is limited, the diversity aspect makes the score less reliable. Therefore, we modified the IS by omitting its diversity evaluation, replacing the mean distribution with a uniform distribution. Specifically, we set $q_i$ in IS to be $1/N$, making the IS of an image $\sum_ip_i\log(Np_i)$, where $N$ is the inception class count and $p_i$ is the predicted probability for the $i_{th}$ class. We denote this modified metric as Quality-only IS (QIS). For the CLIP score, we calculated the mean score between each generated view and the provided text-prompt or image-prompt.

In Tab.\ref{tab:synthesis_quality}, we present comparative results. SD-XL, reflecting the score of test images, achieved the highest QIS and CLIP scores. MVDream, listed as a benchmark for final 3D model quality, shows improved synthesized image quality after 3D fusion due to multi-view consistency. In contrast, Zero123 and Zero123-XL experienced a drop in image quality post-3D fusion due to diffusion inconsistency. Magic123 enhanced the CLIP score over Zero123 by integrating a joint diffusion model. SyncDreamer's quality declined as it diffuses only 16 fixed views, complicating reconstruction. In ImageDream, we evaluated three models for ablation: one with a global controller, another with a local controller, and the last incorporating both local and pixel controllers (Sec.\ref{subsec:multi-level}). ImageDream maintained high image quality in both diffusion and post-3D fusion stages. The local controller, in particular, provided better image CLIP scores post-fusion, thanks to richer image feature representations. The pixel controller model excelled in image CLIP scores during both stages. Notably, ImageDream-pixel ranked second in other scores, with Zero123-XL using a significantly larger dataset (Objaverse-XL~\cite{deitke2023objaversexl}).

However, these scores don't fully encapsulate important aspects like multi-view consistency and geometric correctness. For instance, as Fig.~\ref{fig:quali} demonstrates, zero123-XL, despite having high IS due to easy image classification, showed poorer consistency. Thus, while these scores offer some reliability when consistency is high, future research should aim to develop more comprehensive metrics that accurately capture geometric correctness to better compare different generation algorithms.



\begin{figure}[t]
    \centering\vspace{-0.5em}
    \includegraphics[width=1.01\linewidth]{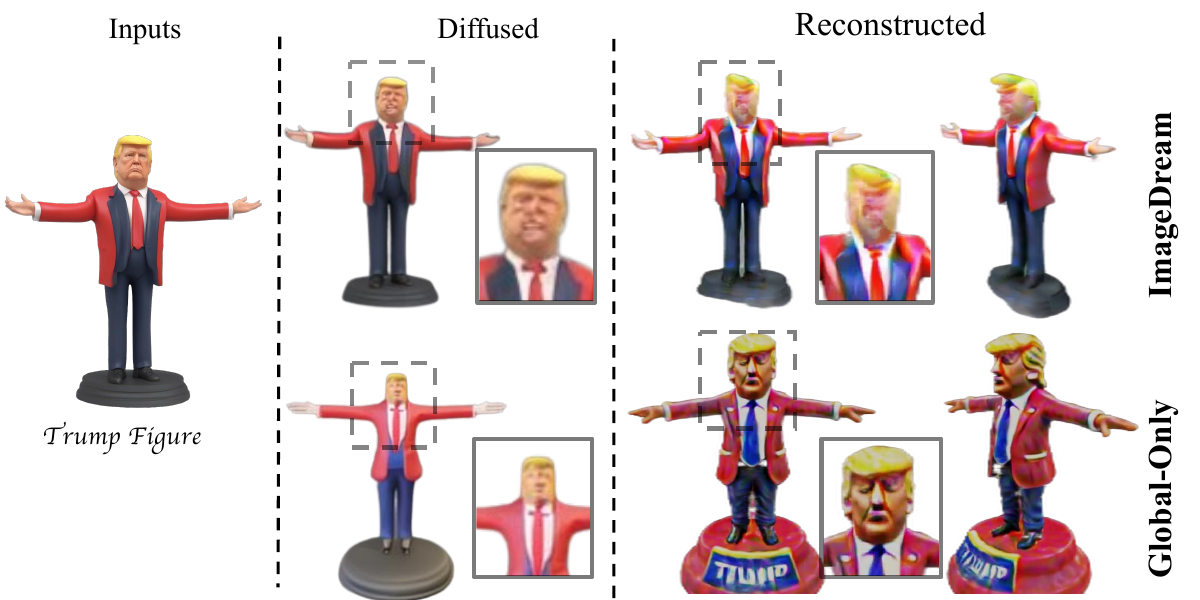}
    \caption{Illustration of a failure case. Where Trumps' face turns blurry after 3D fusion reconstruction (first row). However, the face textures can be generated with model using global-only controller thanks to its semantic global representation.} \vspace{-1.5em}
    \label{fig:faliure}
\end{figure}

\subsection{Limitations}
While the model incorporating the pixel controller achieves the best scores, we observed certain trade-offs, particularly when the image constraints are overly stringent. For instance, in cases like the small facial details of a full-body avatar (shown in Fig.\ref{fig:faliure}), the model struggles to capture these nuances, whereas global control might recover the face based on the text prompt. To address this, as outlined in Sec.\ref{subsec:sds}, the pixel controller model needs to better estimate image intrinsic and extrinsic properties, or a better balance tuning inside of multi-level controllers. This may be solved by exploring the use of larger models, such as SDXL~\cite{podell2023sdxl}, which be our future work.





\section{Conclusion}
We introduce ImageDream, an advanced image-prompt 3D generation model utilizing multi-view diffusion. This model innovatively applies canonical camera coordination and multi-level image-prompt controllers, enhancing control and addressing geometric inaccuracies seen in prior methods. Future improvements could focus on increasing randomness in image-prompts during training to further reduce texture blurriness in the generated models. These steps are expected to further advance the capabilities and applications of ImageDream in 3D model generation.


\newpage
\section{Acknowledgements and Ethics Statement}
We thank our 3D group members of \href{https://likojack.github.io/kejieli/#/home}{Kejie Li}, and our intern \href{https://zeyuan-chen.com/}{Zeyuan Chen} in joint meeting discussion and setup baselines of SyncDreamer, which help complete this paper. In addition, note that the models proposed in this paper aims to facilitate the 3D generation task that is widely demanded in  industry for ethical purpose. It could be potentially applied to unwanted scenarios such as generating violent and sexual content by third-party fine-tuning. Built upon the Stable Diffusion model~\citep{LatentDiffusion}, it might also inherit the biases and limitations to generate unwanted results. Therefore, we believe that the images or models synthesized using our approach should be carefully examined and be presented as synthetic. Such generative models may also have the potential to displace creative workers via automation. That being said, these tools may also enable growth and improve accessibility for the creative industry.
{
    \small
    \bibliographystyle{ieeenat_fullname}
    \bibliography{main}

\begin{thebibliography}{53}
\providecommand{\natexlab}[1]{#1}
\providecommand{\url}[1]{\texttt{#1}}
\expandafter\ifx\csname urlstyle\endcsname\relax
  \providecommand{\doi}[1]{doi: #1}\else
  \providecommand{\doi}{doi: \begingroup \urlstyle{rm}\Url}\fi

\bibitem[SDX()]{SDXL}
stable-diffusion-xl-base-1.0.
\newblock \url{https://huggingface.co/stabilityai/stable-diffusion-xl-base-1.0}.
\newblock Accessed: 2023-08-29.

\bibitem[sdv()]{sdvariation}
Stable diffusion image variation.
\newblock \url{https://huggingface.co/spaces/lambdalabs/stable-diffusion-image-variations}.

\bibitem[Chan et~al.(2022)Chan, Lin, Chan, Nagano, Pan, De~Mello, Gallo, Guibas, Tremblay, Khamis, et~al.]{chan2022efficient}
Eric~R Chan, Connor~Z Lin, Matthew~A Chan, Koki Nagano, Boxiao Pan, Shalini De~Mello, Orazio Gallo, Leonidas~J Guibas, Jonathan Tremblay, Sameh Khamis, et~al.
\newblock Efficient geometry-aware 3d generative adversarial networks.
\newblock In \emph{CVPR}, 2022.

\bibitem[Chan et~al.(2023)Chan, Nagano, Chan, Bergman, Park, Levy, Aittala, Mello, Karras, and Wetzstein]{chan2023genvs}
Eric~R. Chan, Koki Nagano, Matthew~A. Chan, Alexander~W. Bergman, Jeong~Joon Park, Axel Levy, Miika Aittala, Shalini~De Mello, Tero Karras, and Gordon Wetzstein.
\newblock {GeNVS}: Generative novel view synthesis with {3D}-aware diffusion models.
\newblock In \emph{arXiv}, 2023.

\bibitem[Chang et~al.(2015)Chang, Funkhouser, Guibas, Hanrahan, Huang, Li, Savarese, Savva, Song, Su, et~al.]{chang2015shapenet}
Angel~X Chang, Thomas Funkhouser, Leonidas Guibas, Pat Hanrahan, Qixing Huang, Zimo Li, Silvio Savarese, Manolis Savva, Shuran Song, Hao Su, et~al.
\newblock Shapenet: An information-rich 3d model repository.
\newblock \emph{arXiv preprint arXiv:1512.03012}, 2015.

\bibitem[Chen et~al.(2023)Chen, Chen, Jiao, and Jia]{chen2023fantasia3d}
Rui Chen, Yongwei Chen, Ningxin Jiao, and Kui Jia.
\newblock Fantasia3d: Disentangling geometry and appearance for high-quality text-to-3d content creation.
\newblock \emph{arXiv:2303.13873}, 2023.

\bibitem[Deitke et~al.(2023{\natexlab{a}})Deitke, Liu, Wallingford, Ngo, Michel, Kusupati, Fan, Laforte, Voleti, Gadre, VanderBilt, Kembhavi, Vondrick, Gkioxari, Ehsani, Schmidt, and Farhadi]{deitke2023objaversexl}
Matt Deitke, Ruoshi Liu, Matthew Wallingford, Huong Ngo, Oscar Michel, Aditya Kusupati, Alan Fan, Christian Laforte, Vikram Voleti, Samir~Yitzhak Gadre, Eli VanderBilt, Aniruddha Kembhavi, Carl Vondrick, Georgia Gkioxari, Kiana Ehsani, Ludwig Schmidt, and Ali Farhadi.
\newblock Objaverse-xl: A universe of 10m+ 3d objects.
\newblock 2023{\natexlab{a}}.

\bibitem[Deitke et~al.(2023{\natexlab{b}})Deitke, Schwenk, Salvador, Weihs, Michel, VanderBilt, Schmidt, Ehsani, Kembhavi, and Farhadi]{deitke2023objaverse}
Matt Deitke, Dustin Schwenk, Jordi Salvador, Luca Weihs, Oscar Michel, Eli VanderBilt, Ludwig Schmidt, Kiana Ehsani, Aniruddha Kembhavi, and Ali Farhadi.
\newblock Objaverse: A universe of annotated 3d objects.
\newblock In \emph{CVPR}, pages 13142--13153, 2023{\natexlab{b}}.

\bibitem[Deng et~al.(2022)Deng, Yang, Xiang, and Tong]{deng2022gram}
Yu Deng, Jiaolong Yang, Jianfeng Xiang, and Xin Tong.
\newblock Gram: Generative radiance manifolds for 3d-aware image generation.
\newblock In \emph{CVPR}, pages 10673--10683, 2022.

\bibitem[Gao et~al.(2022)Gao, Shen, Wang, Chen, Yin, Li, Litany, Gojcic, and Fidler]{gao2022get3d}
Jun Gao, Tianchang Shen, Zian Wang, Wenzheng Chen, Kangxue Yin, Daiqing Li, Or Litany, Zan Gojcic, and Sanja Fidler.
\newblock Get3d: A generative model of high quality 3d textured shapes learned from images.
\newblock \emph{NeurIPS}, 2022.

\bibitem[Henderson and Ferrari(2020)]{henderson2020learning}
Paul Henderson and Vittorio Ferrari.
\newblock Learning single-image 3d reconstruction by generative modelling of shape, pose and shading.
\newblock \emph{International Journal of Computer Vision}, 2020.

\bibitem[Henderson et~al.(2020)Henderson, Tsiminaki, and Lampert]{henderson2020leveraging}
Paul Henderson, Vagia Tsiminaki, and Christoph~H Lampert.
\newblock Leveraging 2d data to learn textured 3d mesh generation.
\newblock In \emph{CVPR}, 2020.

\bibitem[Huang et~al.(2023)Huang, Wang, Shi, Qi, Zha, and Zhang]{huang2023dreamtime}
Yukun Huang, Jianan Wang, Yukai Shi, Xianbiao Qi, Zheng-Jun Zha, and Lei Zhang.
\newblock Dreamtime: An improved optimization strategy for text-to-3d content creation.
\newblock \emph{arXiv:2306.12422}, 2023.

\bibitem[Jun and Nichol(2023)]{jun2023shap}
Heewoo Jun and Alex Nichol.
\newblock Shap-e: Generating conditional 3d implicit functions.
\newblock \emph{arXiv:2305.02463}, 2023.

\bibitem[Karnewar et~al.(2023)Karnewar, Vedaldi, Novotny, and Mitra]{karnewar2023holodiffusion}
Animesh Karnewar, Andrea Vedaldi, David Novotny, and Niloy~J Mitra.
\newblock Holodiffusion: Training a 3d diffusion model using 2d images.
\newblock In \emph{CVPR}, 2023.

\bibitem[Kingma and Welling(2014)]{kingma2013auto}
Diederik~P Kingma and Max Welling.
\newblock Auto-encoding variational bayes.
\newblock In \emph{ICLR}, 2014.

\bibitem[Lin et~al.(2023{\natexlab{a}})Lin, Gao, Tang, Takikawa, Zeng, Huang, Kreis, Fidler, Liu, and Lin]{lin2023magic3d}
Chen-Hsuan Lin, Jun Gao, Luming Tang, Towaki Takikawa, Xiaohui Zeng, Xun Huang, Karsten Kreis, Sanja Fidler, Ming-Yu Liu, and Tsung-Yi Lin.
\newblock Magic3d: High-resolution text-to-3d content creation.
\newblock In \emph{CVPR}, 2023{\natexlab{a}}.

\bibitem[Lin et~al.(2023{\natexlab{b}})Lin, Liu, Li, and Yang]{lin2023common}
Shanchuan Lin, Bingchen Liu, Jiashi Li, and Xiao Yang.
\newblock Common diffusion noise schedules and sample steps are flawed.
\newblock \emph{arXiv:2305.08891}, 2023{\natexlab{b}}.

\bibitem[Liu et~al.(2023{\natexlab{a}})Liu, Wu, Van~Hoorick, Tokmakov, Zakharov, and Vondrick]{zero123}
Ruoshi Liu, Rundi Wu, Basile Van~Hoorick, Pavel Tokmakov, Sergey Zakharov, and Carl Vondrick.
\newblock Zero-1-to-3: Zero-shot one image to 3d object.
\newblock \emph{arXiv:2303.11328}, 2023{\natexlab{a}}.

\bibitem[Liu et~al.(2023{\natexlab{b}})Liu, Lin, Zeng, Long, Liu, Komura, and Wang]{liu2023syncdreamer}
Yuan Liu, Cheng Lin, Zijiao Zeng, Xiaoxiao Long, Lingjie Liu, Taku Komura, and Wenping Wang.
\newblock Syncdreamer: Learning to generate multiview-consistent images from a single-view image.
\newblock \emph{arXiv preprint arXiv:2309.03453}, 2023{\natexlab{b}}.

\bibitem[Long et~al.(2023)Long, Guo, Lin, Liu, Dou, Liu, Ma, Zhang, Habermann, Theobalt, and Wang]{long2023wonder3d}
Xiaoxiao Long, Yuan-Chen Guo, Cheng Lin, Yuan Liu, Zhiyang Dou, Lingjie Liu, Yuexin Ma, Song-Hai Zhang, Marc Habermann, Christian Theobalt, and Wenping Wang.
\newblock Wonder3d: Single image to 3d using cross-domain diffusion, 2023.

\bibitem[Melas-Kyriazi et~al.(2023)Melas-Kyriazi, Laina, Rupprecht, and Vedaldi]{melas2023realfusion}
Luke Melas-Kyriazi, Iro Laina, Christian Rupprecht, and Andrea Vedaldi.
\newblock Realfusion: 360deg reconstruction of any object from a single image.
\newblock In \emph{CVPR}, 2023.

\bibitem[Mildenhall et~al.(2021)Mildenhall, Srinivasan, Tancik, Barron, Ramamoorthi, and Ng]{mildenhall2021nerf}
Ben Mildenhall, Pratul~P Srinivasan, Matthew Tancik, Jonathan~T Barron, Ravi Ramamoorthi, and Ren Ng.
\newblock Nerf: Representing scenes as neural radiance fields for view synthesis.
\newblock In \emph{ECCV}, 2021.

\bibitem[Nguyen-Phuoc et~al.(2019)Nguyen-Phuoc, Li, Theis, Richardt, and Yang]{nguyen2019hologan}
Thu Nguyen-Phuoc, Chuan Li, Lucas Theis, Christian Richardt, and Yong-Liang Yang.
\newblock Hologan: Unsupervised learning of 3d representations from natural images.
\newblock In \emph{Proceedings of the IEEE/CVF International Conference on Computer Vision}, 2019.

\bibitem[Nguyen-Phuoc et~al.(2020)Nguyen-Phuoc, Richardt, Mai, Yang, and Mitra]{nguyen2020blockgan}
Thu~H Nguyen-Phuoc, Christian Richardt, Long Mai, Yongliang Yang, and Niloy Mitra.
\newblock Blockgan: Learning 3d object-aware scene representations from unlabelled images.
\newblock \emph{NeurIPS}, 2020.

\bibitem[Nichol et~al.(2022)Nichol, Jun, Dhariwal, Mishkin, and Chen]{nichol2022point}
Alex Nichol, Heewoo Jun, Prafulla Dhariwal, Pamela Mishkin, and Mark Chen.
\newblock Point-e: A system for generating 3d point clouds from complex prompts.
\newblock \emph{arXiv:2212.08751}, 2022.

\bibitem[Niemeyer and Geiger(2021)]{niemeyer2021giraffe}
Michael Niemeyer and Andreas Geiger.
\newblock Giraffe: Representing scenes as compositional generative neural feature fields.
\newblock In \emph{CVPR}, 2021.

\bibitem[Niemeyer et~al.(2022)Niemeyer, Barron, Mildenhall, Sajjadi, Geiger, and Radwan]{DBLP:conf/cvpr/NiemeyerBMS0R22}
Michael Niemeyer, Jonathan~T. Barron, Ben Mildenhall, Mehdi S.~M. Sajjadi, Andreas Geiger, and Noha Radwan.
\newblock Regnerf: Regularizing neural radiance fields for view synthesis from sparse inputs.
\newblock In \emph{CVPR}, 2022.

\bibitem[Podell et~al.(2023)Podell, English, Lacey, Blattmann, Dockhorn, M{\"u}ller, Penna, and Rombach]{podell2023sdxl}
Dustin Podell, Zion English, Kyle Lacey, Andreas Blattmann, Tim Dockhorn, Jonas M{\"u}ller, Joe Penna, and Robin Rombach.
\newblock Sdxl: Improving latent diffusion models for high-resolution image synthesis.
\newblock \emph{arXiv:2307.01952}, 2023.

\bibitem[Poole et~al.(2023)Poole, Jain, Barron, and Mildenhall]{poole2022dreamfusion}
Ben Poole, Ajay Jain, Jonathan~T. Barron, and Ben Mildenhall.
\newblock Dreamfusion: Text-to-3d using 2d diffusion.
\newblock In \emph{ICLR}, 2023.

\bibitem[Qian et~al.(2023)Qian, Mai, Hamdi, Ren, Siarohin, Li, Lee, Skorokhodov, Wonka, Tulyakov, and Ghanem]{qian2023magic123}
Guocheng Qian, Jinjie Mai, Abdullah Hamdi, Jian Ren, Aliaksandr Siarohin, Bing Li, Hsin-Ying Lee, Ivan Skorokhodov, Peter Wonka, Sergey Tulyakov, and Bernard Ghanem.
\newblock Magic123: One image to high-quality 3d object generation using both 2d and 3d diffusion priors.
\newblock \emph{arXiv preprint arXiv:2306.17843}, 2023.

\bibitem[Radford et~al.(2021)Radford, Kim, Hallacy, Ramesh, Goh, Agarwal, Sastry, Askell, Mishkin, Clark, et~al.]{CLIP}
Alec Radford, Jong~Wook Kim, Chris Hallacy, Aditya Ramesh, Gabriel Goh, Sandhini Agarwal, Girish Sastry, Amanda Askell, Pamela Mishkin, Jack Clark, et~al.
\newblock Learning transferable visual models from natural language supervision.
\newblock In \emph{ICML}, 2021.

\bibitem[Rombach et~al.(2022)Rombach, Blattmann, Lorenz, Esser, and Ommer]{LatentDiffusion}
Robin Rombach, Andreas Blattmann, Dominik Lorenz, Patrick Esser, and Bj{\"{o}}rn Ommer.
\newblock High-resolution image synthesis with latent diffusion models.
\newblock In \emph{CVPR}, 2022.

\bibitem[Salimans et~al.(2016)Salimans, Goodfellow, Zaremba, Cheung, Radford, and Chen]{InceptionScore}
Tim Salimans, Ian Goodfellow, Wojciech Zaremba, Vicki Cheung, Alec Radford, and Xi Chen.
\newblock Improved techniques for training gans.
\newblock \emph{NeurIPS}, 2016.

\bibitem[Schuhmann et~al.(2022)Schuhmann, Beaumont, Vencu, Gordon, Wightman, Cherti, Coombes, Katta, Mullis, Wortsman, et~al.]{schuhmann2022laion}
Christoph Schuhmann, Romain Beaumont, Richard Vencu, Cade Gordon, Ross Wightman, Mehdi Cherti, Theo Coombes, Aarush Katta, Clayton Mullis, Mitchell Wortsman, et~al.
\newblock Laion-5b: An open large-scale dataset for training next generation image-text models.
\newblock \emph{NeurIPS}, 2022.

\bibitem[Shi et~al.(2023{\natexlab{a}})Shi, Chen, Zhang, Liu, Xu, Wei, Chen, Zeng, and Su]{shi2023zero123++}
Ruoxi Shi, Hansheng Chen, Zhuoyang Zhang, Minghua Liu, Chao Xu, Xinyue Wei, Linghao Chen, Chong Zeng, and Hao Su.
\newblock Zero123++: a single image to consistent multi-view diffusion base model.
\newblock \emph{arXiv preprint arXiv:2310.15110}, 2023{\natexlab{a}}.

\bibitem[Shi et~al.(2023{\natexlab{b}})Shi, Wang, Ye, Mai, Li, and Yang]{shi2023MVDream}
Yichun Shi, Peng Wang, Jianglong Ye, Long Mai, Kejie Li, and Xiao Yang.
\newblock Mvdream: Multi-view diffusion for 3d generation.
\newblock \emph{arXiv:2308.16512}, 2023{\natexlab{b}}.

\bibitem[Shue et~al.(2023)Shue, Chan, Po, Ankner, Wu, and Wetzstein]{shue2023triplanediffusion}
J~Ryan Shue, Eric~Ryan Chan, Ryan Po, Zachary Ankner, Jiajun Wu, and Gordon Wetzstein.
\newblock 3d neural field generation using triplane diffusion.
\newblock In \emph{CVPR}, 2023.

\bibitem[Sitzmann et~al.(2019)Sitzmann, Zollh{\"o}fer, and Wetzstein]{sitzmann2019scene}
Vincent Sitzmann, Michael Zollh{\"o}fer, and Gordon Wetzstein.
\newblock Scene representation networks: Continuous 3d-structure-aware neural scene representations.
\newblock \emph{NeurIPS}, 32, 2019.

\bibitem[Sun et~al.(2023)Sun, Zhang, Shao, Wang, Liu, Xie, and Liu]{sun2023dreamcraft3d}
Jingxiang Sun, Bo Zhang, Ruizhi Shao, Lizhen Wang, Wen Liu, Zhenda Xie, and Yebin Liu.
\newblock Dreamcraft3d: Hierarchical 3d generation with bootstrapped diffusion prior, 2023.

\bibitem[Szymanowicz et~al.(2023)Szymanowicz, Rupprecht, and Vedaldi]{szymanowicz2023viewset}
Stanislaw Szymanowicz, Christian Rupprecht, and Andrea Vedaldi.
\newblock Viewset diffusion:(0-) image-conditioned 3d generative models from 2d data, 2023.

\bibitem[Tang et~al.(2023)Tang, Wang, Zhang, Zhang, Yi, Ma, and Chen]{tang2023make}
Junshu Tang, Tengfei Wang, Bo Zhang, Ting Zhang, Ran Yi, Lizhuang Ma, and Dong Chen.
\newblock Make-it-3d: High-fidelity 3d creation from a single image with diffusion prior.
\newblock \emph{arXiv:2303.14184}, 2023.

\bibitem[Tsalicoglou et~al.(2023)Tsalicoglou, Manhardt, Tonioni, Niemeyer, and Tombari]{tsalicoglou2023textmesh}
Christina Tsalicoglou, Fabian Manhardt, Alessio Tonioni, Michael Niemeyer, and Federico Tombari.
\newblock Textmesh: Generation of realistic 3d meshes from text prompts.
\newblock \emph{arXiv:2304.12439}, 2023.

\bibitem[Wang et~al.(2023{\natexlab{a}})Wang, Du, Li, Yeh, and Shakhnarovich]{wang2023score}
Haochen Wang, Xiaodan Du, Jiahao Li, Raymond~A Yeh, and Greg Shakhnarovich.
\newblock Score jacobian chaining: Lifting pretrained 2d diffusion models for 3d generation.
\newblock In \emph{CVPR}, 2023{\natexlab{a}}.

\bibitem[Wang et~al.(2023{\natexlab{b}})Wang, Zhang, Zhang, Gu, Bao, Baltrusaitis, Shen, Chen, Wen, Chen, et~al.]{wang2023rodin}
Tengfei Wang, Bo Zhang, Ting Zhang, Shuyang Gu, Jianmin Bao, Tadas Baltrusaitis, Jingjing Shen, Dong Chen, Fang Wen, Qifeng Chen, et~al.
\newblock Rodin: A generative model for sculpting 3d digital avatars using diffusion.
\newblock In \emph{CVPR}, 2023{\natexlab{b}}.

\bibitem[Wang et~al.(2023{\natexlab{c}})Wang, Lu, Wang, Bao, Li, Su, and Zhu]{wang2023prolificdreamer}
Zhengyi Wang, Cheng Lu, Yikai Wang, Fan Bao, Chongxuan Li, Hang Su, and Jun Zhu.
\newblock Prolificdreamer: High-fidelity and diverse text-to-3d generation with variational score distillation.
\newblock \emph{arXiv:2305.16213}, 2023{\natexlab{c}}.

\bibitem[Watson et~al.(2023)Watson, Chan, Martin{-}Brualla, Ho, Tagliasacchi, and Norouzi]{DBLP:conf/iclr/WatsonCMHT023}
Daniel Watson, William Chan, Ricardo Martin{-}Brualla, Jonathan Ho, Andrea Tagliasacchi, and Mohammad Norouzi.
\newblock Novel view synthesis with diffusion models.
\newblock In \emph{ICLR}, 2023.

\bibitem[Wu et~al.(2023)Wu, Johnson, Malik, Feichtenhofer, and Gkioxari]{wu2023mcc}
Chao-Yuan Wu, Justin Johnson, Jitendra Malik, Christoph Feichtenhofer, and Georgia Gkioxari.
\newblock Multiview compressive coding for 3d reconstruction.
\newblock In \emph{CVPR}, 2023.

\bibitem[Xu et~al.(2022{\natexlab{a}})Xu, Jiang, Wang, Fan, Shi, and Wang]{Xu_2022_SinNeRF}
Dejia Xu, Yifan Jiang, Peihao Wang, Zhiwen Fan, Humphrey Shi, and Zhangyang Wang.
\newblock Sinnerf: Training neural radiance fields on complex scenes from a single image.
\newblock 2022{\natexlab{a}}.

\bibitem[Xu et~al.(2022{\natexlab{b}})Xu, Jiang, Wang, Fan, Wang, and Wang]{Xu_2022_neuralLift}
Dejia Xu, Yifan Jiang, Peihao Wang, Zhiwen Fan, Yi Wang, and Zhangyang Wang.
\newblock Neurallift-360: Lifting an in-the-wild 2d photo to a 3d object with 360° views.
\newblock 2022{\natexlab{b}}.

\bibitem[Ye et~al.(2023{\natexlab{a}})Ye, Zhang, Liu, Han, and Yang]{ye2023ip}
Hu Ye, Jun Zhang, Sibo Liu, Xiao Han, and Wei Yang.
\newblock Ip-adapter: Text compatible image prompt adapter for text-to-image diffusion models.
\newblock \emph{arXiv preprint arXiv:2308.06721}, 2023{\natexlab{a}}.

\bibitem[Ye et~al.(2023{\natexlab{b}})Ye, Wang, Li, Shi, and Wang]{ye2023consistent}
Jianglong Ye, Peng Wang, Kejie Li, Yichun Shi, and Heng Wang.
\newblock Consistent-1-to-3: Consistent image to 3d view synthesis via geometry-aware diffusion models.
\newblock \emph{arXiv preprint arXiv:2310.03020}, 2023{\natexlab{b}}.

\bibitem[Zhou and Tulsiani(2023)]{zhou2023sparsefusion}
Zhizhuo Zhou and Shubham Tulsiani.
\newblock Sparsefusion: Distilling view-conditioned diffusion for 3d reconstruction.
\newblock In \emph{CVPR}, 2023.

\end{thebibliography}
}


\end{document}